\documentclass[conference]{IEEEtran}
\IEEEoverridecommandlockouts
\usepackage{cite}
\usepackage{amsmath,amssymb,amsfonts}
\usepackage{algorithmic}
\usepackage{graphicx}
\usepackage{textcomp}
\usepackage{xcolor}
\def\BibTeX{{\rm B\kern-.05em{\sc i\kern-.025em b}\kern-.08em
    T\kern-.1667em\lower.7ex\hbox{E}\kern-.125emX}}

\usepackage{multirow}

\begin{document}

\title{Transformer Neural Networks Attending to Both Sequence and Structure for Protein Prediction Tasks\\
}

\author{\IEEEauthorblockN{Anowarul Kabir}
\IEEEauthorblockA{\textit{Department of Computer Science} \\
\textit{George Mason University}\\
akabir4@gmu.edu}
\and

\IEEEauthorblockN{Amarda Shehu}
\IEEEauthorblockA{\textit{Department of Computer Science}\\
\textit{George Mason University}\\
ashehu@gmu.edu}
}
\maketitle


\begin{abstract}
The increasing number of protein sequences decoded from genomes is opening up new avenues of research on linking protein sequence to function with transformer neural networks. Recent research has shown that the number of known protein sequences supports learning useful, task-agnostic sequence representations via transformers. In this paper, we posit that learning joint sequence-structure representations yields better representations for function-related prediction tasks. We propose a transformer neural network that attends to both sequence and tertiary structure. We show that such joint representations are more powerful than sequence-based representations only, and they yield better performance on superfamily membership across various metrics. 
\end{abstract}
\section{Introduction}
\label{sec:Introduction}

While we have known for decades that the amino-acid sequence of a protein molecule determines to a great extent its activities in the living cell, predicting protein function from sequence remains a hallmark problem in molecular biology~\cite{Dekker_NatNanoTech18}. While throughput technologies have greatly increased the number of protein sequences in public repositories, very few of them have been experimentally characterized. For instance, only about $1$\% of the sequences in the UniProtKB database have been functionally characterized in wet laboratories~\cite{UniProt}. This gap, amplified by the potential for therapeutic applications, continues to motivate computational research on function prediction~\cite{BileschiColwell22}. 

The computational literature is rich for various reasons. First, the term protein function is ill-defined, as it includes a possibly large set of diverse activities of a protein in the cell. The level of detail at describing protein function varies. One can predict superfamily membership, family membership, different levels of the gene ontology (GO) hierarchy, or detailed interactions with specific small molecules or other macromolecules~\cite{ZhaoYu20}. In addition, many of the cellular activities of a protein molecule may be instigated by structural changes, but modeling structural dynamics is by itself an open problem~\cite{MaximovaShehuPCB16}.

The availability of protein databases providing functional characterization, such as Pfam~\cite{Pfam21} and SCOP~\cite{SCOP20}, including other benchmark datasets, has permitted the design of many machine learning algorithms, including deep learning-based approaches. A review of such literature is beyond the scope of this paper, but we direct interested readers to recent surveys in~\cite{ZhaoYu20, VuJung21}.  

The line of research we advance in this paper is on building meaningful representations of protein molecules to support function-related prediction tasks. Research in representation learning is also very rich. Before the wide availability of volumes of sequence data, researchers had to understand how a specific biological activity placed constraints on sequence and/or structure and think deeply about how to encode those constraints in sequence- and/or structure-based function-encoding representations. The rise of big (macromolecular) data is now providing an opportunity to learn such representations directly from the data and possibly in a more general manner beyond a specific biological activity. 

In particular, transformers, which have revolutionized Natural Language Processing, present a unique opportunity for representation learning. Work in this direction has just started. Recent work in~\cite{NambiarBCB20} trains a transformer over protein sequences to learn sequence-based representations that are shown to be powerful in predicting protein family membership. In this paper, we advance this work, leveraging the fact that by pre-training on task-agnostic sequence representations, transformers are highly appealing to support a variety of prediction tasks. 

In this paper we posit that learning joint sequence-structure representations yields better representations for function-related prediction tasks. The main contribution of this paper is the integration of tertiary structure and its encoding in a manner that allows utilizing the attention mechanism. We employ a transformation-invariant representation of tertiary structure through the concept of a contact map, which allows the transformer to learn meaningful sequence-structure representations. 

In essence, the transformer-based model we put forward attends to both the sequence and associated tertiary structure of a protein molecule. Due to the ready availability of tertiary structures for protein molecules already characterized in SCOP, we demonstrate the utility of the joint sequence-structure representation for the task of superfamily prediction. However, the approach can support various prediction tasks; for instance, we envision how AlphaFold2~\cite{AlphaFold2} can be utilized to generate a reasonable tertiary structure of a given amino-acid sequence~\cite{AlphaFold2Database}. The experimental evaluation we describe in this paper shows that the learned sequence-structure representations are more powerful than sequence-based representations alone, and they yield better performance on superfamily membership across various metrics, even on a dataset that is inherently highly imbalanced.

The rest of this paper proceeds as follows. Section~\ref{sec:RelatedWork} briefly summarizes related work and preliminaries. Methodological details are related in Section~\ref{sec:Methods}. The experimental evaluation is detailed in Section~\ref{sec:Results}, and the paper concludes in Section~\ref{sec:Conclusion} with a summary and outlook.
\section{Related Work}
\label{sec:RelatedWork}

\subsection{Deep Neural Networks in Computational Biology}

Deep neural networks are increasingly becoming the methods of choice in protein structure prediction, protein function prediction, genome engineering, systems biology, and phylogenetic inference. In protein structure prediction, we find convolutional neural networks (CNNs), Residual Networks (ResNet), and Transformers. In protein function prediction, we find CNNs, ResNets, Recurrent Neural Networks (RNNs), and Graph Neural Networks (GNNs). In Genome Engineering, Multi-layer Perceptrons (MLPs) and CNNs are popular. In systems biology, we find CNNs, RNNs, and Variational Autoencoders (VAEs). CNNs and ResNets are popular in phylogenetic inference. Protein structure and function predicted are two of the computational biology areas where deep learning has been a major success, and even paradigm shifting in protein structure prediction~\cite{AlphaFold2}. We refer the interested reader to Ref.~\cite{SapovalTreangen22} for a review of deep learning literature in computational biology.

\subsection{Deep Neural Networks for Protein Function Prediction}

Limited and imbalanced datasets, a possibly large space of functions/activities, and the hierarchical nature of  some functional annotations, such as the GO hierarchy, are key challenges for machine learning methods for protein function prediction, including deep learning. To address some of these challenges, methods have leveraged a diversity of features from sequence, structure, interaction networks, biomedical literature, and more. We highlight three recent methods based on deep learning, DeepGO~\cite{Kulmanov2017}, DeepGOPlus~\cite{Kulmanov2019}, and deepNF~\cite{Gligorijevi2018}. DeepGO incorporates a CNN to learn sequence-level embeddings and combines them with knowledge graph embeddings obtained from Protein-Protein Interaction (PPI) networks~\cite{Kulmanov2017}. DeepGOPlus uses convolutional filters of different sizes with individual max-pooling to learn dense feature representations of protein sequences embedded via one-hot encoding~\cite{Kulmanov2019}. The authors show that combining the outputs from CNN with homology-based predictions result in better predictive accuracy. deepNF utilizes a multi-modal denoising auto encoder to extract features from multiple heterogeneous interaction networks and show that the resulting model outperform methods based on matrix factorization and linear regression~\cite{Gligorijevi2018}.

The utilization of transformers for function prediction is in its infancy. The PRoBERTa model in~\cite{NambiarBCB20} is pre-trained to learn task-agnostic sequence representations of amino-acid sequences. Since there is no inherent notion of words in a given amino-acid sequence, the authors in~\cite{NambiarBCB20} restrict the vocabulary size to $10,000$ words and use the byte-pair encoding algorithm~\cite{10.5555/177910.177914} to identify words. The PRoBERTa model in~\cite{NambiarBCB20} is then fine-tuned to solve two prediction tasks, protein family memberships and protein-protein interactions. Work in~\cite{vig2021bertology}, though not focusing on any particular prediction task, analyzes various transformer models, such as BERT, ALBERT, and XLNet, through the lens of attention and shows them capable of capturing proximity constraints among amino acids in a tertiary structure of an uncomplexed/free protein or a protein binding site of a bound protein. 
\section{Preliminaries}
\label{sec:Preliminaries}

Here we overview the concept of attention and the transformer architecture. Readers already familiar with these preliminaries are encouraged to skip ahead to Section~\ref{sec:Methods}.

\subsection{Attention}

In this work, we focus only on the attention mechanism which was first introduced by~\cite{bahdanau2016neural} in neural machine translation tasks. The goal in~\cite{bahdanau2016neural} is to solve fixed-length encoding vector issues that arise when the representation of long and short sequences is forced to be the same. The authors compute the alignment scores for a given encoded hidden state and previous decoder output. This indicates how well the elements of the input sequence align with the current output at a given position. The alignment model is represented as a function, which can be implemented by a feed-forward neural network. Then a softmax function is run over the alignment scores to compute the probability distribution, termed as attention weights. These are multiplied with the hidden state to compute the weighted sum and the context vector as output of the attention layer. 

Work in~\cite{vaswani2017attention} reformulates the attention mechanism using queries ($Q$), keys ($K$), and values ($V$). In this formulation, each query vector ($q$) is matched against a set of keys ($k_i$) by computing the dot product between them: 
\vspace*{-2mm}
\begin{equation}
    \label{eq:attn_step_1}
    e_{q, k_i} = q \cdot  k_i\\[-2mm]
\end{equation}

The softmax function is then run to generate the distribution as weights in \ref{eq:attn_step_2}. Finally, the attention is computed by a weighted sum of the value vectors, and the generated attention weights as in \ref{eq:attn_step_3}.
\vspace*{-2mm}
\begin{equation}
    \label{eq:attn_step_2}
    \alpha_{q, k_i} = softmax(e_{q, k_i})
\end{equation}
\vspace*{-2mm}
\begin{equation}
    \label{eq:attn_step_3}
    attention(q, K, V) = \sum_i \alpha_{q, k_i} v_{k_i}\\[-3mm]
\end{equation}

When the generalized attention mechanism is presented with a sequence of words, it takes the query vector attributed to some specific word in the sequence and scores it against each key in the database. In doing so, it captures how the word under consideration relates to the others in the sequence. Then it scales the values according to the attention weights in order to retain focus on those words that are relevant to the query. In doing so, it produces an attention output and attention weight for the word under consideration.

Work in~\cite{vaswani2017attention} also proposed a novel encoder-decoder architecture without relying on any use of recurrence and convolutions utilizing only the attention mechanism. The encoder maps an input sequence to a sequence of continuous representations. It consists of $N_{enc}$ identical layers where each layer is composed of two sublayers. The generalized attention mechanism and a fully connected feed-forward network are implemented in the first and second sublayer, respectively. 

The decoder receives the output of the encoder together with the decoder output at the previous time step and generates an output sequence and thus it works in an auto-regressive manner. It is composed of $N_{dec}$ identical layers where each layer consists of three sublayers. The first sublayer receives the previous output of the decoder stack, augments it with positional information, and implements multi-head self-attention over it. While the encoder is designed to attend to all words in the input sequence, the decoder is modified to attend only to the preceding words. Hence, the prediction for a word at a position  can only depend on the known outputs for the words that come before it in the sequence. The second layer implements a multi-head self-attention mechanism, which receives the queries from the previous decoder sublayer, and the keys and values from the output of the encoder. This allows the decoder to attend to all of the words in the input sequence. The third layer implements a fully connected feed-forward network. 

Each of the two sublayers have a residual connection around it. In all sublayers, the layer normalization and positional encoding are applied in both the encoder and decoder part. The output of the decoder finally passes through a fully connected layer, followed by a softmax layer, to generate a prediction for the next word of the output sequence. 

Following the vanilla transformer architecture summarized above, the Bidirectional Encoder Representations from Transformers (BERT)~\cite{DBLP:journals/corr/abs-1810-04805} is proposed by focusing on the encoder part only. BERT can be trained in an unsupervised manner for representation learning and then can be fine-tuned on downstream tasks in a supervised fashion. Many variants of BERT are now available, including ALBERT~\cite{DBLP:journals/corr/abs-1909-11942}, RoBERTa~\cite{DBLP:journals/corr/abs-1907-11692} and DistilBERT~\cite{DBLP:journals/corr/abs-1910-01108}.

\section{Methods}
\label{sec:Methods}

We first relate details on the prediction task, input dataset, and then describe the transformer neural network that attends to both sequence and structure.

\subsection{Prediction Task}

For the purpose of evaluation, we focus here on superfamily membership prediction. In summary, a protein family is a group of proteins that share a common evolutionary origin, which is reflected by their related functions and similarities in sequence or structure. A protein superfamily is a large group of distantly-related proteins~\cite{Knudsen2010}. A protein superfamily may contain more than one family.

The Structural Classification of Proteins (SCOP)~\cite{Andreeva2019}, which is created by manual inspection but aided by a variety of automated methods, provides a detailed and comprehensive description of the structural and evolutionary relationships among all proteins whose structure is known. The proteins in this database are additionally annotated with the superfamily to which they belong. Therefore, the SCOP database is 
ideal for our experimental evaluation. 


\subsection{Dataset}
\label{subsec:dataset}

We use the SCOP database to extract $36,535$ proteins. Each SCOP entry contains a link to the tertiary structure available for it in the Protein Data Bank (PDB)~\cite{PDB03}; the PDB is a repository of tertiary structures of protein molecules. Many quality checks need to be performed to prepare the dataset, as we relate below.

\begin{itemize}

\item If a protein tertiary strucure cannot be found in the PDB, the corresponding SCOP entry is removed from our dataset. An example of such an entry is  6qwj, which is now obsolete in the PDB. 

\item If the DSSP~\cite{Kabsch1983,Joosten2010} program fails to compute secondary structure elements (often because the tertiary structure has missing amino acids), the respective entry is removed.

\item If the tertiary structure, in PDB format, does not pass the chain-id test by Biopython~\cite{Cock09}, a library we use to process tertiary structures, the entry is removed. Biopyton allows us to check whether the PDB format passes compliance checks. 

\item If a PDB entry contains one or more missing amino acids (that is, the tertiary structure is incomplete), the corresponding SCOP entry is not included in the dataset.

\end{itemize}

\begin{figure*}[ht]
\centering
\begin{tabular}{c}
\includegraphics[width=0.8\textwidth]{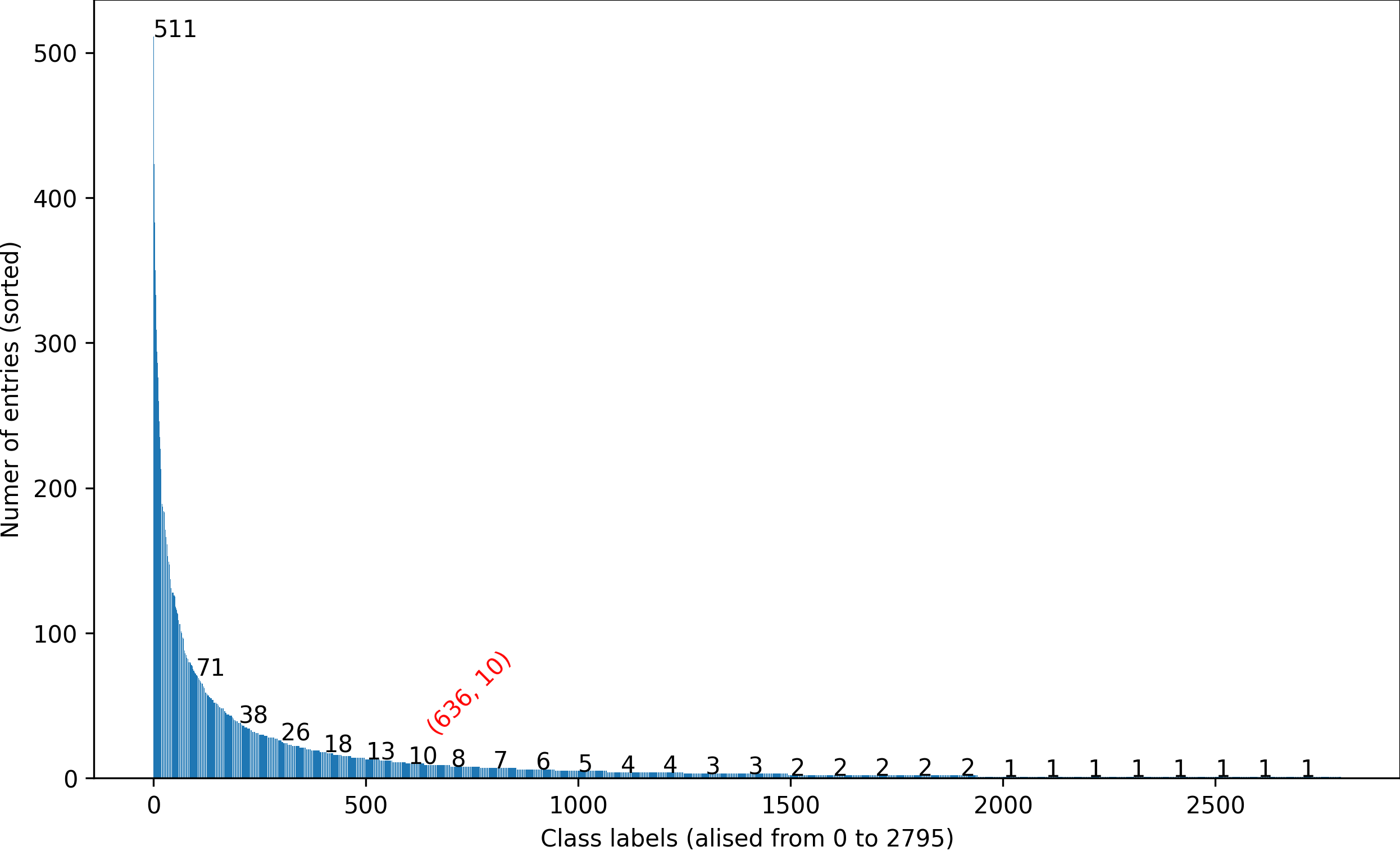}\\[-4mm]
\end{tabular}
\caption{The class distribution of our dataset is shown here. Classes are ordered by their size, highest to lowest. Size is shown on the y-axis and annotated on top of each $100$th bar corresponding to a class. The x-axis relates the classes, which are $0$-indexed. Red font draws attention to the fact that only $637$ of the $2,796$ classes have at least ten members.}
\vspace*{-2mm}
\label{fig:data_distribution}
\end{figure*}

At the end of this process, we have $26,994$ complete entries which belong to $2,796$ superfamily classes. The proteins in the dataset have a maximum length of $1,460$ amino acids. The highly-skewed class distribution over the dataset is shown in Fig.~\ref{fig:data_distribution}, where the classes are sorted/ordered by the number of entries per class (highest to lowest). The size of each class is related on the y-axis and shown on each $100$th bar corresponding to the class. The class labels are aliased from $0$ to $2,795$ numeric entries. Fig.~\ref{fig:data_distribution} also shows that there are only $637$ classes (note that they are $0$-indexed) that have at least $10$ members.

We divide the pre-processed, clean dataset into a standard train, validation and test set. The train dataset is prepared so that it contains $70\%$ instances from each class. From the remaining dataset, $50\%$ of the data instances from each class are selected for the validation set, and the remaining are placed in the test set. Our train, validation and test set contain $24,538$, $4,458$ and $5,862$ data instances, respectively. Following this protocol, we are able to maintain the same class distribution of the dataset throughout the train, validation and test set.

\subsection{Methods}
\label{subsec:method}

We refer to the model we propose as ProToFormer, which stands for \underline{Pro}tein  \underline{To}pological Trans\underline{former}). ProToFormer learns joint sequence-structure representations of protein molecules. We investigate two variants based on the embedding dimension for each amino acid. We include in our experimental evaluation the PRoBERTa~\cite{NambiarBCB20} model (as a baseline model) to obtain a task-agnostic sequence-based representation of a protein molecule and fine-tune it on the superfamily classification task.

We first relate details on ProToFormer and then on the FT-PRoBERTa used as baseline. This section concludes with implementation details, which include values of the various hyperparameters for each model.

\subsubsection{ProToFormer}

We recall that the vanilla transformer encoder~\cite{opennmt} and its variants, such as BERT~\cite{DBLP:journals/corr/abs-1810-04805}, utilize the bidirectional information flow for a given input. As such, the model has the ability to look at every position with the same attention probability. For a specific position $i$, the model applies an attention mechanism from the $0$-th to $l$-th positions of an $l$-length sequence to compute the abstract relation among tokens by minimizing a loss function (commonly, cross-entropy). 

A protein entry in our dataset contains both sequence and tertiary structure information. However, the existing transformer-based model, PRoBERTa, leverages only sequence information. We hypothesize that the topological information encoded in a tertiary structure might provide stronger signals than bidirectional sequence-level signals while applying an attention mechanism for a task at hand. Both protein sequence and structure determine function; as related in Section~\ref{sec:Introduction}, sequence can vary more rapidly, whereas structure is under a stronger evolutionary pressure to maintain the function of a protein molecule. 

To encode the topological information available in a tertiary structure of a protein molecule, we utilize a 2D representation of the tertiary structure, the \emph{contact map}. The contact map is SE3-invariant. Though referred to as a map, it is an $N \times N$ symmetric Boolean matrix encoding the threshold-binarized spatial proximities of the $N$ amino acids of a protein in a given tertiary structure. Spatial proximity is measured through the Euclidean distance over the central carbon (CA) atoms that are found in each amino acid. The typical threshold used is $8$\AA. Distances above these thresholds are encoded as $0$ in the contact map; that is, no contact. Otherwise, the entries are filled with $1$'s, indicating contact between corresponding amino acids. In summary, the $(i,j)$ entry in a contact map corresponding to a tertiary structure indicates that there is a contact between the $i$-th and $j$-th amino acids if the Euclidean distance between them does not exceed $8.0$\AA.

\begin{figure}[ht]
\centering
\begin{tabular}{c}
\textbf{(a) Amino-acid sequence}\\
\includegraphics[width=\columnwidth]{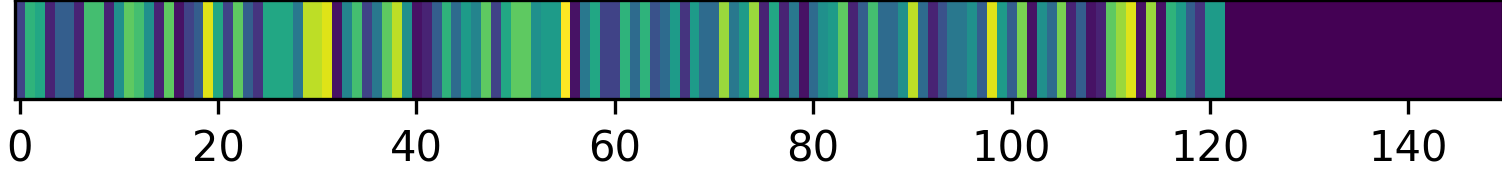}\\
\textbf{(b) Key-padding mask}\\
\includegraphics[width=\columnwidth]{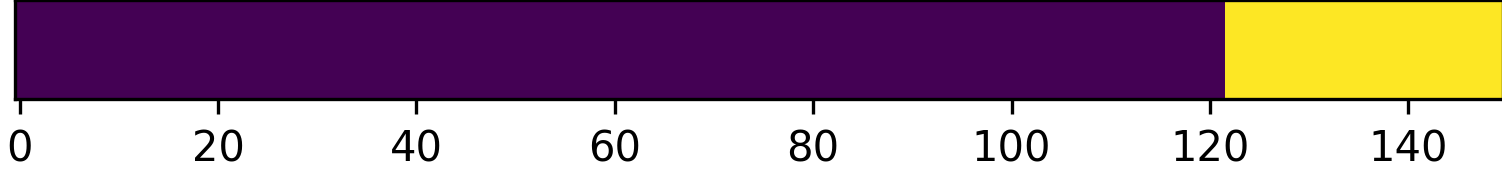}\\
\textbf{(c) Attention mask}\\
\hspace*{-3mm}
\includegraphics[width=\columnwidth]{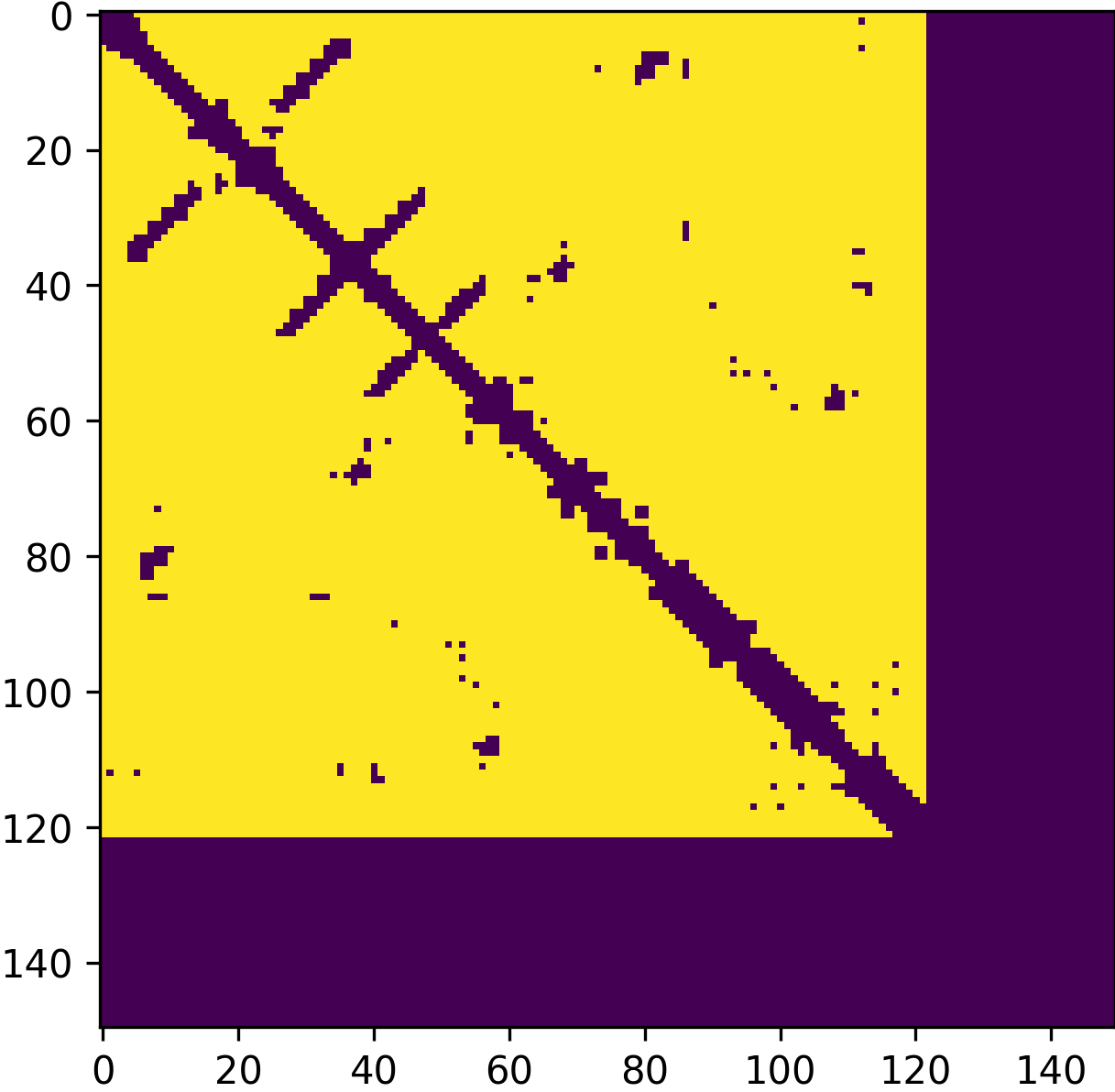}\\[-3mm]
\end{tabular}
\caption{(a) Each amino acid is encoded as a $1$ to $20$ numeric number, inclusive, and zero is kept as a padding index. (b) Corresponding key-padding mask with a padded region (colored as yellow at the right). (c) Respective attention mask, where the yellow region indicates $i$-th amino acid (row) not to attend $j$-th amino acid (column). The padded region of the attention mask will be nullified by the key-padding mask.}
\label{fig:model_input}
\end{figure}

We input the contact map to the transformer encoder as the attention mask to provide available information about the topology. Thus, the input of the model consists of three items: the protein sequence, key-padding mask, and the attention mask. In Fig.~\ref{fig:model_input}, we show an example of an amino-acid sequence and its corresponding key-padding mask and attention mask for PDB-ID: 3h8d, chain-id: C, residue (amino acid)-id range: $1,143-1,264$ of sequence length $122$; this is padded to length $150$ in the illustration.

Note that unlike PRoBERTa ProToFormer does not need to figure out what the analogous of words are in a sequence. In essence, a single amino-acid is a word, and this allows the model to pay attention to semantic relationships between amino acids as encoded in the tertiary structure. In this manner, ProToFormer learns single amino-acid representations in a higher-dimensional space that encodes structural constraints. We consider two variants of ProToFormer based on the embedding dimension. In our experimental evaluation, we experiment with an embedding dimension of $256$ versus $128$. We refer to these two models as ProToFormer (128-SEQ+CM) and ProToFormer (256-SEQ+CM), where SEQ+CM refers to the utilization of both sequence and contact map. A third model is included for comparison, ProToFormer (128-SEQ), which does not attend to contacts but to sequence alone. Column 2 in Table~\ref{table:hyperparameters} relates architectural details and fine-tuning hyperparameters. 

\subsubsection{FT-PRoBERTa}

We employ a fine-tuned PRoBERTa as a baseline model against which to compare PRoToFormer and its variants. We recall that PRoBERTa learns sequence-only representations. PRoBERTa~\cite{NambiarBCB20} is trained on a masked language modeling (MLM) task on a corpus of $450K$ unique sequences taken from UniProtKB/Swiss-Prot~\cite{UniProt}. To the model provided in~\cite{NambiarBCB20}, we add a classification layer and train the resulting model, FT-PRoBERTa (FT for fine-tuned) on the training dataset related above for the superfamily classification task. The architectural details and fine-tuning hyperparameters of FT-PRoBERta can be found in Table~\ref{table:hyperparameters} (Column $3$). 

\subsubsection{Implementation Details}
\label{subsubsection:ImplementationDetails}

In ProToFormer we utilize the vanilla transformer encoder architecture~\cite{opennmt} and apply the PyTorch multi-head attention implementation. We consider each amino acid as a word and apply a word embedding layer of dimension $128$ or $256$. We find that when we utilize $256$ the model converges fast and performs better, but we relate results on both settings in Section~\ref{sec:Results}. We note that the maximum length of a sequence is set at $512$. 

We apply the Adam optimizer with weight-decay as $0.01$ and cross-entropy loss to train the model. Since the class distribution is highly skewed, we compute the class weights using Scikit-learn's \textit{compute-class-weight} function. The output layer is implemented as taking the last layer's embedding of an input data and outputs a probability distribution over the $2,796$ classes. We use the mini-batch algorithm of batch size $64$ and $128$ to train and validate each model. The best model is archived at the best validation loss. All the hyperparameters are shown in Table~\ref{table:hyperparameters} (Column $2$). The hyperparameters of the best model are shown in boldface font.
 
\begin{table}[t]
\caption{Architectural details and hyperparameter values of ProToFormer and FT-PRoBERTa.}
\label{table:hyperparameters}
\vspace*{-2mm}
\begin{center}
\begin{small}
\begin{sc}
\begin{tabular}{lccr}
\textbf{Hyperparams} & \textbf{ProToFormer} & \textbf{FT-PRoBERTa} \\
Max-length   & 512 & 512 \\
Embed-dim  & 128, \textbf{256}  & 768 \\
\#-Attn-heads  & 8 & 12 \\
\#-Enc-layers  & 5 & 5 \\
Dropout  & 0.1 & 0.1 \\
LR  & 1e-3, \textbf{1e-4}, 1e-5 & 1e-4, \textbf{1e-5} \\
Batch size  & \textbf{64}, 128 & 32 \\
Attention  & Contact-map (CM)  &  N/A \\  
\#-Weights & 13M & 44M \\
\end{tabular}
\end{sc}
\end{small}
\end{center}
\vskip -0.1in
\end{table}

\section{Results}
\label{sec:Results}

We first compare all four models, ProToFormer (128-SEQ+CM), ProToFormer (256-SEQ+CM), ProToFormer (128-SEQ), and FT-PRoBERTa on along accuracy, precision, recall, F1-score, and AUC-ROC. Since the class distribution over the dataset is skewed, we utilize the weighted precision, recall, and F1-score by the support set and with configuring metrics as outputting $1$ for a class if the support set is $0$. The AUC-ROC is computed for each instance, and we report the average. 

Table~\ref{table:metrics} reports the accuracy (ACC), precision (PRE), recall (REC), F1-score (F1), and AUC-ROC score (AR) of all four models. The comparison shows that ProToFormer (128-SEQ), the ProToFormer that attends to sequence alone, performs similarly to  FT-PRoBERTa on all performance metrics. The two better-performing models along all metrics are ProToFormer (256-SEQ+CM) and ProToFormer (128-SEQ), with ProToFormer (256-SEQ+CM) achieving the best performance. Table~\ref{table:metrics} makes the case that adding the contact map to the training process gives the model an opportunity to learn structural information and improve performance by at least $\sim20\%$ on all performance metrics. Further, increasing the embedding dimension from $128$ to $256$ results in improvements of at least $\sim2\%$ on all performance metrics. 

\begin{table}[htbp]
\caption{Performance comparison among the four models along accuracy (ACC), precision (PRE), recall (REC), F1-score (F1), and AUC-ROC score (AR). The consecutive two rows for each model show performance on a metric on the validation and test set, respectively. The highest value on a metric on each set is highlighted in boldface font.}
\label{table:metrics}
\vskip 0.15in
\begin{center}
\begin{small}
\begin{sc}
\begin{tabular}{p{7em}p{2em}p{2em}p{2em}p{2em}p{2em}p{2em}}
\textbf{Model} & \textbf{Acc} & \textbf{Pre} & \textbf{Rec} & \textbf{F1} & \textbf{AR} \\

\multirow{3}{2.4cm}{FT-PRoBERTa}
& 0.542 & 0.600 & 0.542 & 0.523 & 0.989 \\
& 0.472 & 0.614 & 0.472 & 0.432 & 0.966 \\\\ \smallskip

\multirow{3}{2.4cm}{ProToFormer (128-SEQ)} 
& 0.526 & 0.600 & 0.526 & 0.526 & 0.987 \\
& 0.486 & 0.578 & 0.486 & 0.461 & 0.984 \\\\ \smallskip

\multirow{3}{2.4cm}{ProToFormer (128-SEQ+CM)} 
& 0.720 & 0.762 & 0.720 & 0.713 & 0.992 \\
& 0.664 & 0.741 & 0.664 & \textbf{0.664} & \textbf{0.985} \\\\ \smallskip

\multirow{3}{2.4cm}{ProToFormer (256-SEQ+CM)} 
& \textbf{0.742} & \textbf{0.783} & \textbf{0.742} & \textbf{0.734} & \textbf{0.992} \\
& \textbf{0.678} & \textbf{0.752} & \textbf{0.678} & 0.643 & 0.979  \\[1mm]                   

\end{tabular}
\end{sc}
\end{small}
\end{center}
\vskip -0.1in
\end{table}

\begin{figure*}[htbp]
\centering
\begin{tabular}{c}
\includegraphics[width=0.51\textwidth]{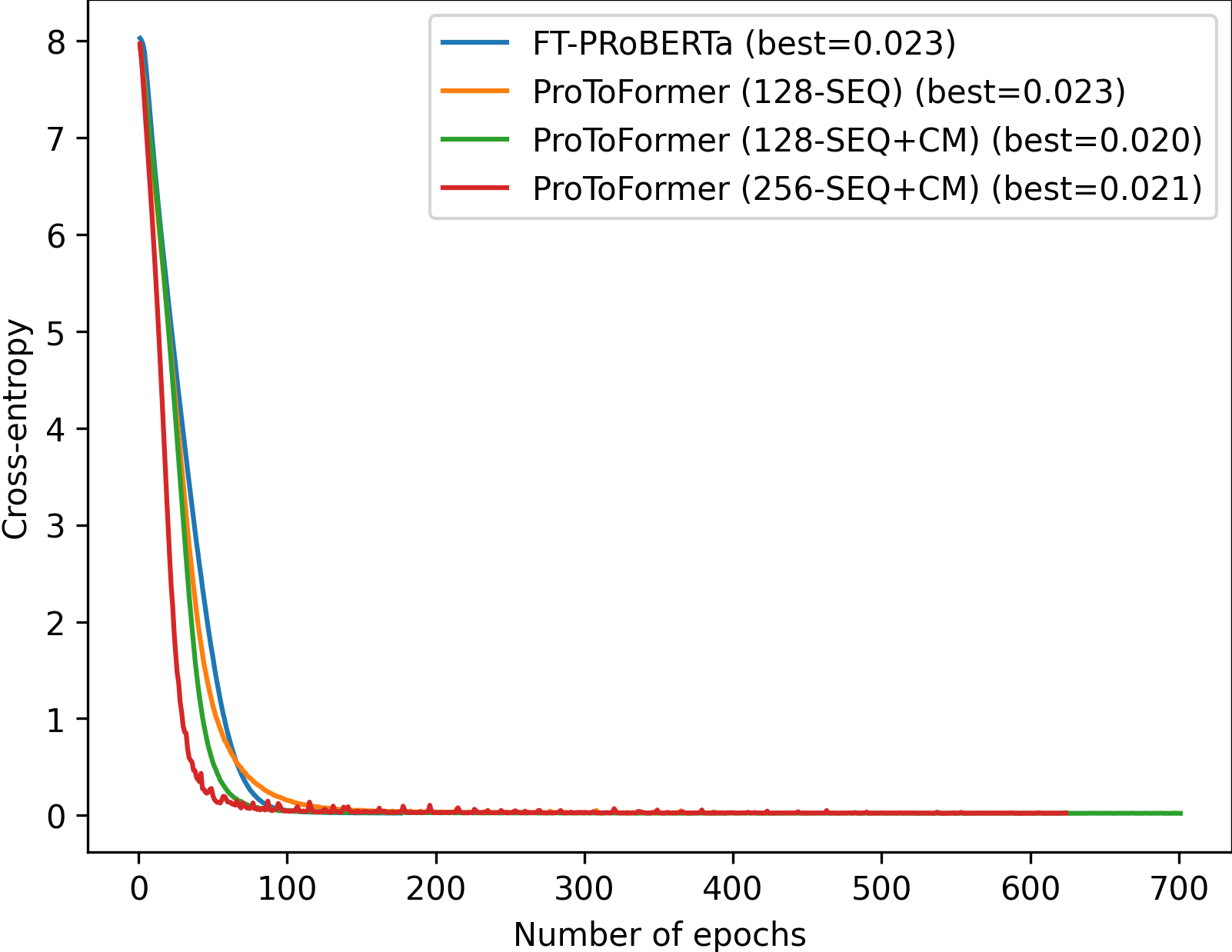}
\end{tabular}
\hspace*{-2mm}
\begin{tabular}{cc}
\includegraphics[width=0.51\textwidth]{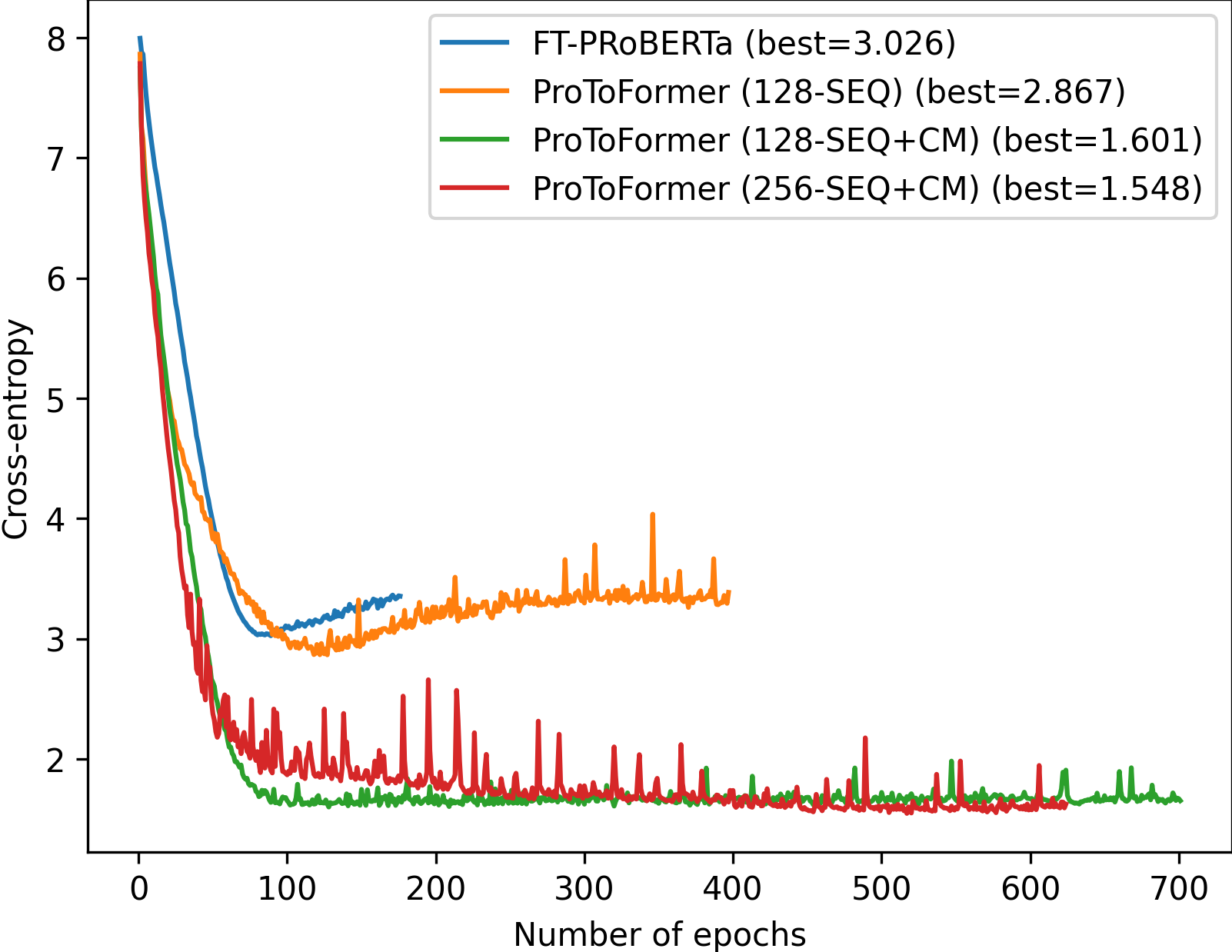} & \hspace*{-3mm}
\includegraphics[width=0.52\textwidth]{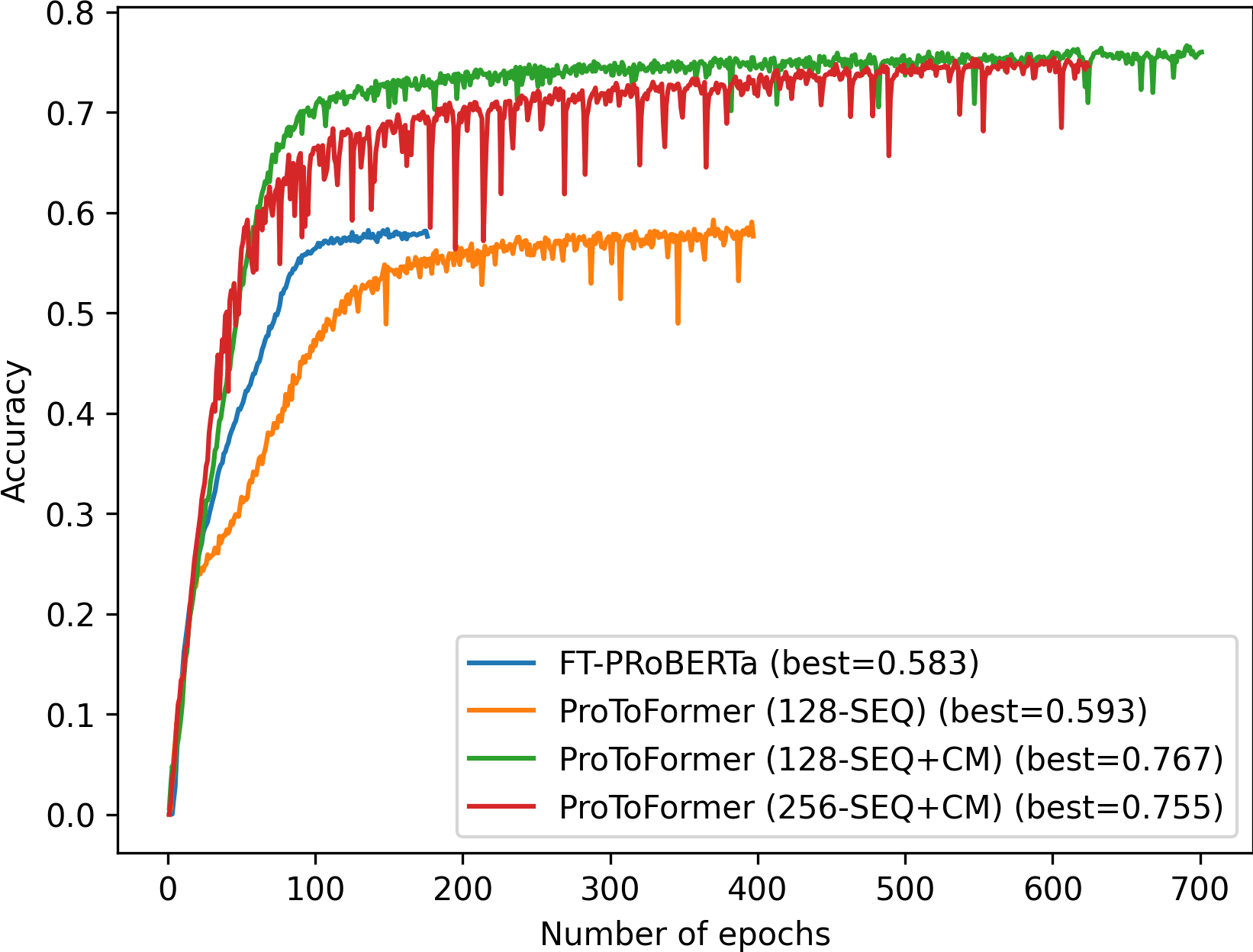}
\end{tabular}
\caption{Top panel: Cross-entropy loss on the training set is tracked over the training epochs for each model. Bottom panel left: Cross-entropy loss on the validation set is tracked over the training epochs for each of the models. Bottom panel right: Accuracy on the validation set is tracked over the training epochs for each of the models.}
\label{fig:PerformanceOverEpochs}
\end{figure*}

Fig.~\ref{fig:PerformanceOverEpochs} provides a more detailed view of the models during training. The top panel shows the cross-entropy loss and accuracy ($y$-axis) over the number of epochs ($x$-axis) for the training for each of the four models. The bottom panel shows the validation loss and accuracy, respectively, over the number of training epochs for the four models. We note that no special attempt is made to fine-tune the models. In addition, the number of trainable parameters on our ProToFormer models is approximately lower by $3$ magnitudes over FT-PRoBERTa.

The above results support the conclusion that ProToFormer (256-SEQ+CM) is the best model. We now investigate in greater detail the performance of this model on four categories of  classes, those with less than $10$ members, those with at least $10$ members, those with less than $30$ members, and those with at least $30$ members. Table~\ref{table:datam_th} relates the performance in each case for both the validation and test set along the various metrics: ACC, PRE, REC, and F1.

Table~\ref{table:datam_th} shows that on classes with at least $10$ data points, the model improves the performance by $\sim1\%$ for both the test and validation set over classes with fewer than $10$ members. If we increase the threshold (th) from $10$ to $30$, the performance improves by at least $4\%$ in all metrics. 

\begin{table}[htbp]
\caption{Performance analysis of ProToFormer (256-SEQ+CM) on classes with less than $10$ members, at least $10$ members, less than $30$ members, and at least $30$ members.}
\label{table:datam_th}
\vskip 0.15in
\begin{center}
\begin{small}
\begin{sc}
\begin{tabular}{lccccccr}

\textbf{Data} & \textbf{th} & \textbf{Acc} & \textbf{Pre} & \textbf{Rec} & \textbf{F1} \\

\multirow{2}{1.5cm}{Val}  & $\ge$10  & 0.753 & 0.795 & 0.753 & 0.753 \\
                        & $<$10   & 0.598 & 0.969 & 0.598 & 0.597 \\
                        \\
\multirow{2}{1.5cm}{Test} & $\ge$10  & 0.752 & 0.791 & 0.752 & 0.750 \\
                        & $<$10   & 0.436 & 0.931 & 0.436 & 0.426 \\
                        \\
\multirow{2}{1.5cm}{Val}  & $\ge$30  & 0.777 & 0.841 & 0.777 & 0.798 \\
                        & $<$30   & 0.642 & 0.894 & 0.642 & 0.661 \\
                        \\
\multirow{2}{1.5cm}{Test} & $\ge$30  & 0.783 & 0.843 & 0.783 & 0.801 \\
                        & $<$30   & 0.529 & 0.850 & 0.529 & 0.522 \\

\end{tabular}
\end{sc}
\end{small}
\end{center}
\vskip -0.1in
\end{table}

\begin{figure}[ht]
\centering
\begin{tabular}{c}
\includegraphics[width=0.49\textwidth]{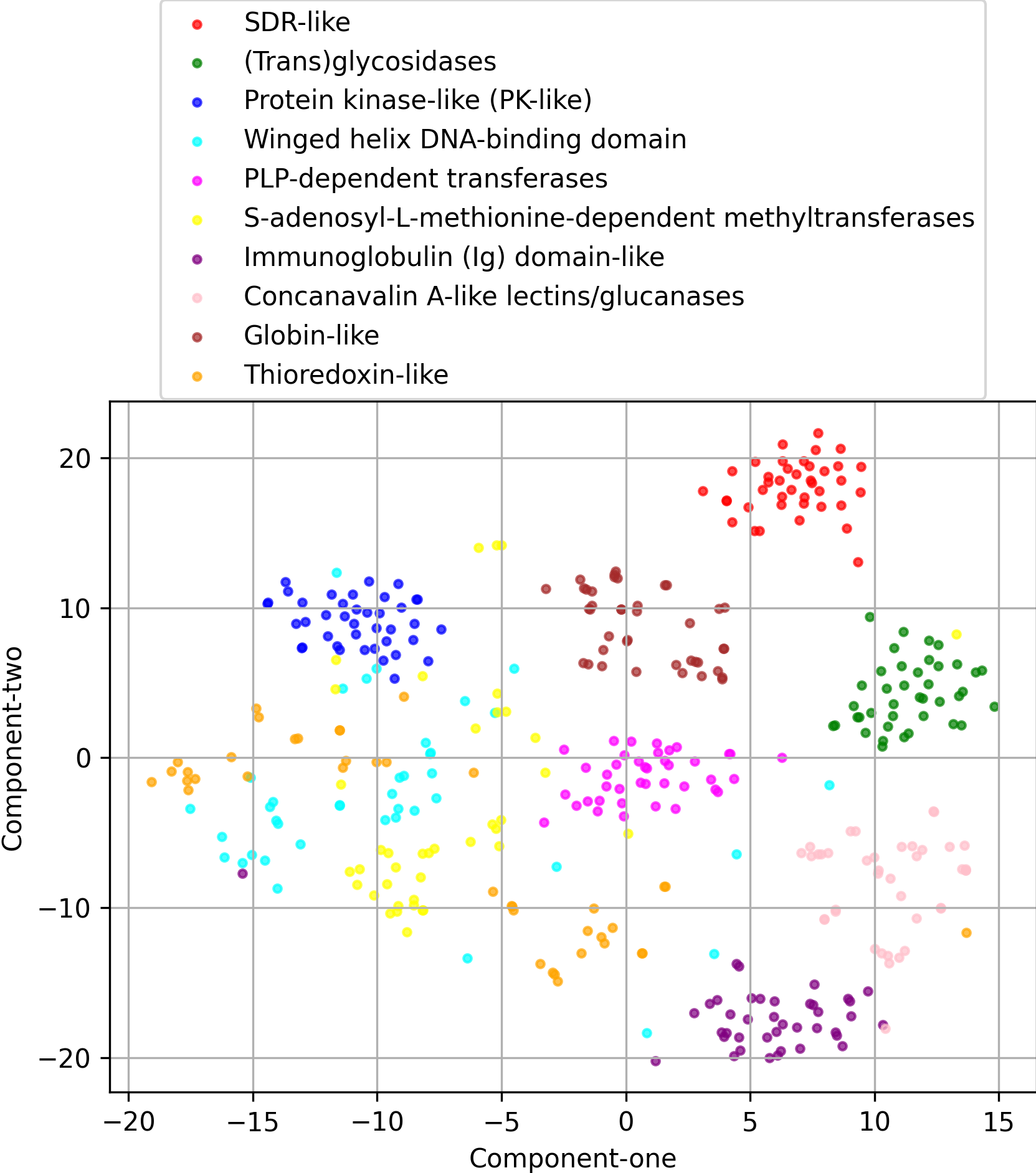}\\[-3mm]
\end{tabular}
\caption{The learned space is projected onto two components via t-SNE. The projections are color-coded by superfamily class. The data shown is restricted to $10$ classes drawn at random among those with at least $40$ members.}
\label{fig:EmbeddingVisualization}
\end{figure}

Finally, in Fig.~\ref{fig:EmbeddingVisualization} we visualize the joint sequence-structure embedding space learned by PRoToFormer (256-SEQ+CM). We map the last layer features on two components via t-SNE~\cite{Maaten08} so we can obtain sequence embeddings in two dimensions. Projections are color-coded by superfamily classes. For ease of visualization, we restrict to $10$ classes drawn at random among those with at least $40$ members. Fig.~\ref{fig:EmbeddingVisualization} shows good separation among the classes and co-localization of the sequence+structure-function space, providing further support for the summary performance related above.

\section{Conclusion}
\label{sec:Conclusion}

The increasing number of protein sequences decoded from genomes is opening up new avenues of research on linking protein sequence to function with deep neural networks. In particular, by pre-training on task-agnostic sequence representations, transformers are becoming increasingly appealing to support a variety of prediction tasks. 

In this paper, we advance budding research on adapting transformers for the protein sequence universe by utilizing them to learn joint sequence-structure representations. The foundation of this line of work rests of the central role of structure and our decades-long knowledge that function places evolutionary pressure on both sequence and structure (and even stronger pressure on structure). The novelty of the work we propose is in utilizing the attention mechanism in a transformer to attend to both sequence and tertiary structure; the latter is encoded as a contact map, which in turn encodes semantic constraints to attend to during the task-agnostic pre-training. We evaluate the learned representations on the superfamily membership classification task.  The experimental evaluation clearly shows that the joint sequence-structure representation confers higher performance on this task than sequence alone. 

We hope the venue of work we have started in this paper opens up further research directions. We note that one of the reasons we focus on the superfamily membership prediction task here is due to the ready availability of tertiary structures of the training data. However, one can envision utilizing AlphaFold2, for instance, to generate tertiary structure information to leverage for joint sequence-structure representations and expand the prediction tasks to other dimensions of protein function.

\section*{Acknowledgment}
This work was supported in part by the National Science Foundation under Grant No. 1900061.

\bibliography{refs/AKabirShehu_ICML2022,refs/Shehu_Refs}
\bibliographystyle{IEEEtran}

\end{document}